\documentclass[journal]{IEEEtran}

\ifCLASSINFOpdf
  \usepackage[pdftex]{graphicx}
\else
\fi
\usepackage{amsmath}
\usepackage{tabularx}
\usepackage{array}
\usepackage{algorithmic}
\usepackage{algorithm}

\usepackage{array}

\usepackage[caption=false,font=normalsize,labelfont=sf,textfont=sf]{subfig}

\usepackage{gensymb}
\usepackage{amssymb}

\begin{document}
%
\title{Dual Encoder-Decoder based Generative Adversarial Networks for Disentangled Facial Representation Learning}

\author{Cong~Hu,
        Zhen-Hua Feng,~\IEEEmembership{Member,~IEEE},
        Xiao-Jun~Wu,
        Josef Kittler,~\IEEEmembership{Life~Member,~IEEE}
\IEEEcompsocitemizethanks{\IEEEcompsocthanksitem Cong Hu and Xiao-Jun Wu are with the School of Internet of Things Engineering, Jiangnan University, Wuxi 214122, Jiangsu Province, China, also with the Jiangsu Provincial Laboratory of Pattern Recognition and
Computational Intelligence, Jiangnan University, Wuxi 214122, Jiangsu Province, China. (Corresponding author: Xiao-Jun Wu)
\protect\\
E-mail: wxhucong@163.com; wu\_xiaojun@jiangnan.edu.cn
\IEEEcompsocthanksitem Zhen-Hua Feng and Josef Kittler are with the Centre for Vision, Speech and Signal Processing, University of Surrey, Guildford GU2 7XH, UK.
\protect\\
E-mail: z.feng@surrey.ac.uk; j.kittler@surrey.ac.uk}
\thanks{}}

%
%

\markboth{Journal of \LaTeX\ Class Files,~Vol.~14, No.~8, August~2015}%
{Shell \MakeLowercase{\textit{\textit{et al.}}}: Bare Demo of IEEEtran.cls for IEEE Journals}
%



\maketitle

\begin{abstract}
To learn disentangled representations of facial images, we present a Dual Encoder-Decoder based Generative Adversarial Network (DED-GAN). In the proposed method, both the generator and discriminator are designed with deep encoder-decoder architectures as their backbones. To be more specific, the encoder-decoder structured generator is used to learn a pose disentangled face representation, and the encoder-decoder structured discriminator is tasked to perform real/fake classification, face reconstruction, determining identity and estimating face pose. We further improve the proposed network architecture by minimising the additional pixel-wise loss defined by the Wasserstein distance at the output of the discriminator so that the adversarial framework can be better trained. Additionally, we consider face pose variation to be continuous, rather than discrete in existing literature, to inject richer pose information into our model. The pose estimation task is formulated as a regression problem, which helps to disentangle identity information from pose variations. The proposed network is evaluated on the tasks of pose-invariant face recognition (PIFR) and face synthesis across poses. An extensive quantitative and qualitative evaluation carried out on several controlled and in-the-wild benchmarking datasets demonstrates the superiority of the proposed DED-GAN method over the state-of-the-art approaches. 
\end{abstract}

\begin{IEEEkeywords}
Disentangled representation learning, encoder-decoder, generative adversarial networks, face synthesis, pose invariant face recognition.
\end{IEEEkeywords}

%
\IEEEpeerreviewmaketitle

\section{Introduction}
Benefiting from the rapid development of deep learning and the easy access to a large number of annotated face images, face recognition~\cite{hu2015face} has advanced significantly in recent years. Although impressive performance has been achieved on several benchmarking databases, pose variation is still one of the crucial bottlenecks for many practical applications~\cite{song2018dictionary}. Facial appearance variations caused by poses are even larger than those caused by different identities~\cite{kan2014stacked}. 
To mitigate this difficulty, a number of approaches have been proposed for pose-invariant face recognition (PIFR). 
Existing PIFR methods can be divided into three categories. 
One approach is to remap non-frontal faces to frontal ones, and then extract facial features from frontalised faces for better face representation~\cite{hassner2015effective,sagonas2015robust,yim2015rotating,kittler20163d, koppen2018gaussian}. 
The second one is to learn pose-invariant representations directly from non-frontal faces~\cite{chan2017face,masi2016pose,schroff2015facenet,feng2016unified}. 
The last category aims to learn disentangled facial representations so that identity-preserving features can be disentangled from pose variation~\cite{peng2017reconstruction,tran2018representation}. Our proposed method belongs to the last category.

The consensus regarding desirable properties of good representations of data has recently been established in~\cite{bengio2013representation,lake2017building,hu2018semi,hu2018discriminative}. Disentanglement, one of the characteristic properties of good representation, is a kind of distributed feature representation in which disjoint dimensions of a latent code reflect different high-level generative factors of data. The disentanglement is also often described as statistical independence; each independent factor is expected to be semantically well aligned with the human intuition regarding the data generative factors. Specifically, the disentangled representation can separate explanatory factors that interact non-linearly in real-world data, such as object shapes, material properties, light sources and so on. A representation distilling each important factor of data into a single independent direction is hard to learn, but it is highly valuable for many other downstream tasks like PIFR and face synthesis across views~\cite{ridgeway2016survey,higgins2017scan}.

Deep generative models facilitate learning disentangled representations. It is a methodology that enables learning the probability distribution of data and generating new samples according to control codes in a latent space. By learning the appropriate parameters, deep generative models can generate new data mimicking the distribution of the target data. Once a disentangled representation is learned, the disjoint dimensions of the hidden code model the data generative factors separately. These underlying factors have the potential to explain the major variations in the data. When only one factor varies but all others are fixed, the generated sequence of samples can show an interpretable change to human beings. For example, when we generate a hand-written digit, a component of the code may be associated with the stroke width. When its value is changed, the stroke width of the generated digit becomes smaller. In recent years, Variational Auto-Encoder (VAE)~\cite{kingma2013auto} and Generative Adversarial Networks (GAN)~\cite{goodfellow2014generative} based methods as two notable branches of deep generative model have successfully been used in the disentangled representation learning. For instance, $\beta$-VAE~\cite{higgins2017beta} learns disentangled latent codes by encouraging the latent distribution to be close to the standard normal distribution, in which each random variable is independent. DC-IGN~\cite{kulkarni2015deep} is another VAE-based generative model for disentangled representation learning. However, DC-IGN may not be applicable to  unstructured in-the-wild images, since it achieves disentanglement by providing batch training samples with one attribute being fixed. InfoGAN~\cite{chen2016infogan} also uses statistical independence, which is motivated by the principle of maximization of mutual information. The Disentangled Representation learning GAN (DR-GAN)~\cite{tran2018representation} learns generative and discriminative facial representations, which disentangle the face identity from pose so that it can better handle cross-pose recognition. DR-GAN is also similar to the prior work~\cite{yim2015rotating} in which joint representation learning and face rotation are explored with a multi-task CNN. In summary, most of the existing works disentangle the factors by using statistical independence of a prior distribution. 

Although DR-GAN has achieved impressive performance in face synthesis across poses and PIFR, it has some problems: 1) The process of training of DR-GAN is not stable. In a few stable cases, a mode collapse often occurs, producing degenerate images; 2) The pose variations are categorised into several distinct classes by a one hot vector. Consequently, although it is a strong prior, the pose information is insufficient for disentangled facial representation learning. In order to improve the training stability of GAN, the encoder-decoder structured discriminator has been successfully used in EBGAN~\cite{zhao2016energy} and BEGAN~\cite{berthelot2017began}, which is also used as a backbone network in our method. To achieve stable model training, an equilibrium enforcing method was proposed in BEGAN, in which a hyper-parameter is introduced to balance the generator and discriminator during the model training. Different from the classical GANs, BEGAN aims to match the auto-encoder loss distributions, not the between sample distributions. We also introduce an equilibrium enforcing strategy in our method. However, in contrast to BEGAN, our method not only matches the distributions between samples like in typical GANs, but also the distributions of the reconstruction losses of samples, which is conducive to better representation learning. Accordingly, pixel-wise reconstruction error is used as another loss function, aside identity loss and pose estimation in our GAN model. 

DR-GAN codes the pose into several classes with a one-hot vector, incurring information loss in the process. In fact pose changes continuously, non-linearly but smoothly. For this reason we represent pose code by a continuous variable rather than in a discrete form. This also allows to estimate  pose by regression rather than classification.

This paper addresses the problem of learning a generative model for disentangled facial representation extraction. By combining the advanced techniques of GAN-based representation learning methods, we propose to learn disentangled pose-robust features by modelling the complex non-linear transform between face images with different poses through a dual encoder-decoder structured deep neural network in an adversarial way, namely Dual Encoder-Decoder based Generative Adversarial Networks (DED-GAN). The proposed network is evaluated in terms of the quality of face synthesis of different views on the one hand and pose-invariant face recognition (PIFR) on the other hand. Our contributions are summarised as follows.
\begin{itemize}
    \item A new GAN architecture with fast and stable convergence is proposed for disentangled facial representation learning.
    
    \item Our proposed method can generate a face with arbitrary pose variations. 
    
    \item The proposed method learns identity-preserving features simultaneously.
    
    \item To the best of our knowledge, this is the first attempt to use pose regression for disentangled face representation. The proposed continuous pose variation model provides more detailed information about the pose. It is used explicitly to control the manifold of identity-preserving face synthesis.
    
    \item Experiments in PIFR and face synthesis across poses demonstrate the advantage of our method on multiple benchmarking databases.
\end{itemize}

The rest of the paper is organised as follows: We first overview the existing literature related to the proposed method in Section~\ref{sec_related_work}. Then we present the proposed DED-GAN in Section~\ref{sec_proposed} and introduce the implementation details in Section~\ref{sec_implementation}. An ablation study and experimental results are reported in Section~\ref{sec_experiment}. Last, the conclusion is drawn in Section~\ref{sec_conclusion}.  
 
\section{Related work}
\label{sec_related_work}
\subsection{Generative Adversarial Network}
Recently, the state of the art in deep generative models, especially in VAE~\cite{kingma2013auto} and GAN~\cite{goodfellow2014generative}, have advanced significantly. 
As one of the most promising deep neural networks, GAN has attracted widespread attention from the computer vision and machine learning communities.
It provides a simple, yet powerful way to estimate data distribution and generate realistic samples by the zero-sum two-player game~\cite{denton2015deep}. 
Through modelling a real sample distribution, a GAN can encourage the generated samples to move towards the true image manifold, and thus generate photo-realistic images with plausible high-frequency details. 
However, the classical GAN suffers from computational problems, \textit{e.g.} the inferior performance caused by unbalanced training of the generator without a comparable attention given to updating the discriminator.
A collapsed generator will lose the capacity to fit the target data distribution.
To address the aforementioned model collapse issue, some improved GAN architectures have been proposed.
For example, Zhao \textit{et al.} ~\cite{zhao2016energy} proposed energy-based GAN (EBGAN) that considers the generator and discriminator as energy functions. 
Salimans \textit{et al.}~\cite{salimans2016improved}  introduced a bag of tricks to address GAN training strategies and achieved great performance on semi-supervised learning.
Karras \textit{et al.}~\cite{karras2017progressive} used a strategy of progressively growing the generator and discriminator of a GAN for improved image generation quality, stability, and variation. 
Further, Arjovsky \textit{et al.}~\cite{arjovsky2017wasserstein} presented Wasserstein GAN (WGAN) using the earth mover's distance. 
They proved that WGAN is able to avoid the mode collapse problem to a certain extent. 

Existing GAN models can handle most of the challenging cases, in which the pose, illumination and expression of faces are unconstrained. 
For example, Radford \textit{et al.}~\cite{radford2015unsupervised} designed DC-GAN that evaluates a set of constraints on the architectural topology of convolutional GANs, which make the model stable to train. 
Huang \textit{et al.}~\cite{huang2017beyond} focused on the local patches that have some semantic meaning and proposed TP-GAN. 
Li \textit{et al.}~\cite{li2017generative} focused on the missing parts of the face and came up with a novel two adversarial losses as well as a semantic parsing loss to complete the faces. He \textit{et al.}~\cite{he2019attgan} edited the face images with desired attributes while preserving other details by encoder-decoder structured GAN. Both~\cite{gauthier2014conditional} and \cite{lu2017conditional} applied an extension of GAN to a conditional setting and showed their utility in many tasks, including image in-painting~\cite{pathak2016context}, super-resolution~\cite{ledig2017photo}, style transfer~\cite{li2016combining}, face attribute manipulation~\cite{shen2017learning} and even data augmentation for classification models~\cite{shrivastava2017learning,zheng2017unlabeled}. The VariGAN model was proposed by Zhao \textit{et al.}~\cite{zhao2017multi} to solve the problem of generating multi-view images from a single viewpoint. 
Tran \textit{et al.}~\cite{tran2017disentangled} put forward DR-GAN, which fuses the pose information and is able to take one or multiple face images with yaw angles as input to achieve pose invariant facial representation learning. 
Similarly, Antipov \textit{et al.}~\cite{antipov2017face} concentrated on improving face synthesis in cross-age scenarios. 
Considering scene structure and context, Yang \textit{et al.}~\cite{yang2017lr} presented LR-GAN that learns generated image background and foreground separately and recursively to produce a completely natural or face image. 

These successful GANs provide a strong motivation to learn disentangled facial representation and to develop a method for different view synthesis. However, there are several crucial issues with GANs such as training being unstable and a quantitative evaluation proving difficult. The previous work either focuses on the stability of training, the task of synthesising images, or using the features in the discriminator for image recognition. In contrast, we propose an innovative method for constructing the generator for disentangled representation learning, which is stable. The proposed DED-GAN method is also quantitatively evaluated for pose invariant face recognition.

\subsection{Pose Invariant Representation Learning}
In conventional face recognition methods, local descriptors~\cite{chan2013multiscale,daugman1985uncertainty,ahonen2006face,dalal2005histograms} and metric learning~\cite{chen2013blessing,weinberger2009distance} are often used to tackle the effect of pose variation. 
In contrast, deep learning methods handle pose variation through building pose-specific or pose-agnostic models with specific loss functions~\cite{chen2009ranking,wen2016discriminative}. 
For instance, the DeepFace~\cite{taigman2014deepface} model uses a deep CNN coupled with 3D face alignment. 
The inception architecture, utilised in FaceNet~\cite{schroff2015facenet}, which in turn is used in DeepID2+~\cite{sun2015deeply} and DeepID3~\cite{sun2015deepid3} where multi-task learning and metric learning are performed simultaneously. 
However, such data-driven methods heavily rely on well-annotated data. Collecting labelled data covering all variations is time-consuming and labour-intensive. 
Our proposed Dual Encoder-Decoder based GAN (DED-GAN) presents an idea similar to Disentangled Representation learning GAN (DR-GAN)~\cite{tran2018representation}, which considers both face rotation and representation learning in a unified network. 
However, our proposed model differs from DR-GAN in the following aspects:
1) we use a continuous pose code for disentangling face representation in DED-GAN, as it provides more detailed information about the pose as a strong prior for training. 2) DR-GAN suffers from poor generalisation and from optimisation difficulties, which limit its effectiveness in face synthesis and face recognition. In contrast, our DED-GAN overcomes these issues by disentangling the pose by means of pose regression and adding face reconstruction as a side task. 

\section{The proposed approach}
\label{sec_proposed}
\begin{figure*}[t]
\centering
\subfloat[Vanilla GAN]{
\label{Vanilla GAN}
  \includegraphics[trim = 50mm 15mm 50mm 10mm, clip, height = .4\linewidth]{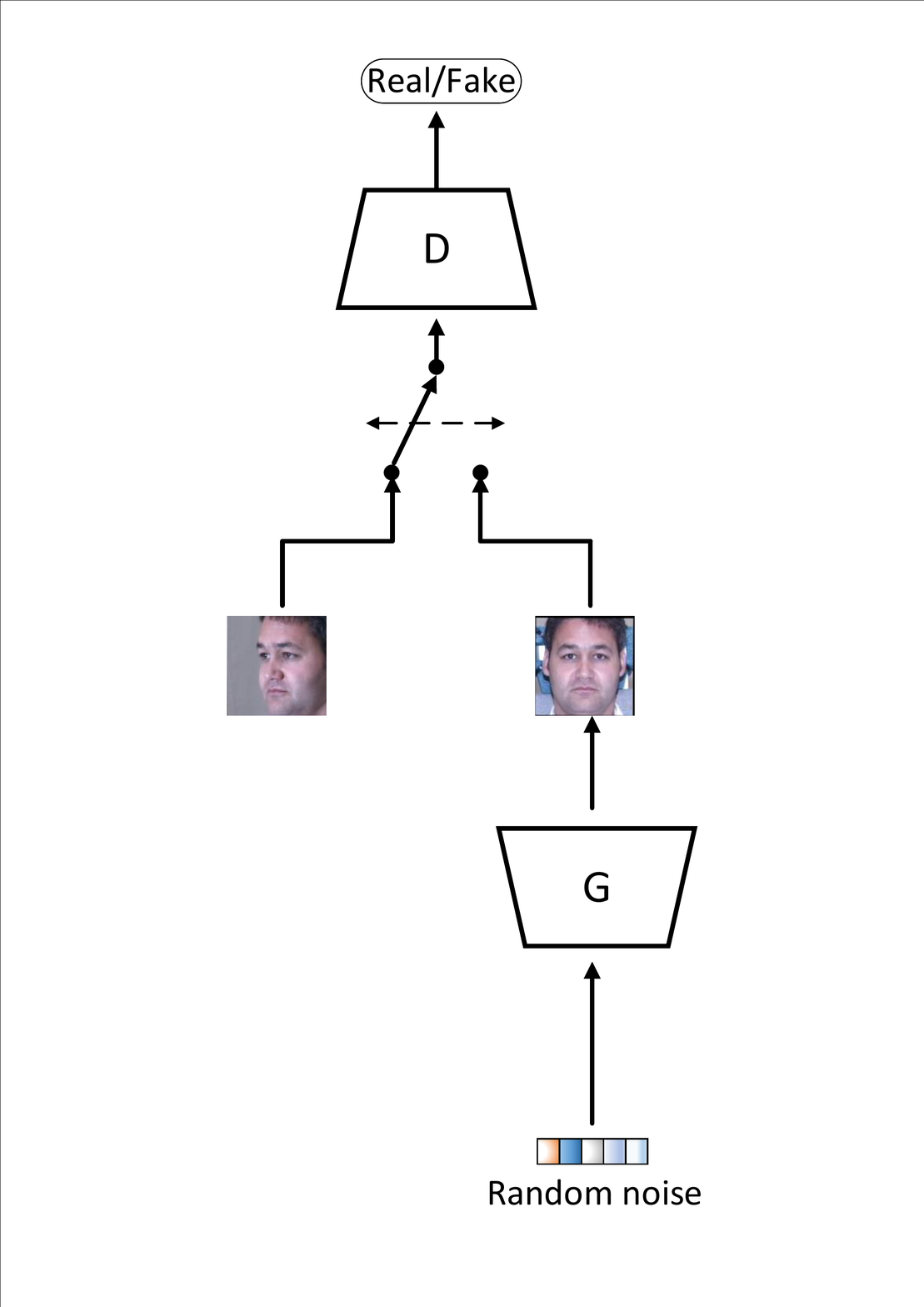}
}
\subfloat[Auxiliary Classifier GAN]{
 \label{Auxiliary Classifier GAN}
  \includegraphics[trim = 10mm 5mm 10mm 10mm, clip, height = .41\linewidth]{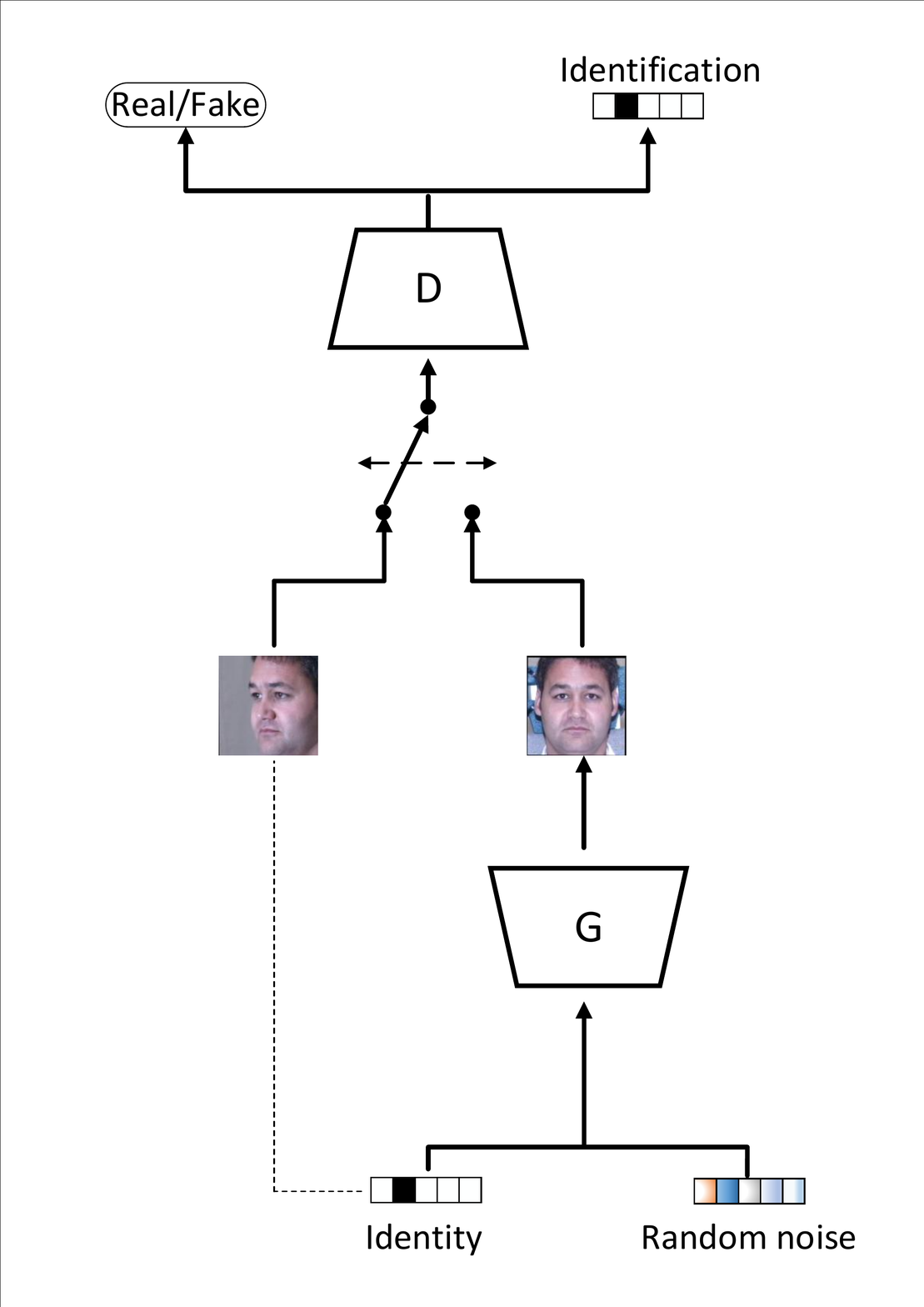}
}
\subfloat[DR-GAN]{
 \label{DR-GAN}
  \includegraphics[trim = 10mm 1mm 10mm 260mm, clip, height = .44\linewidth]{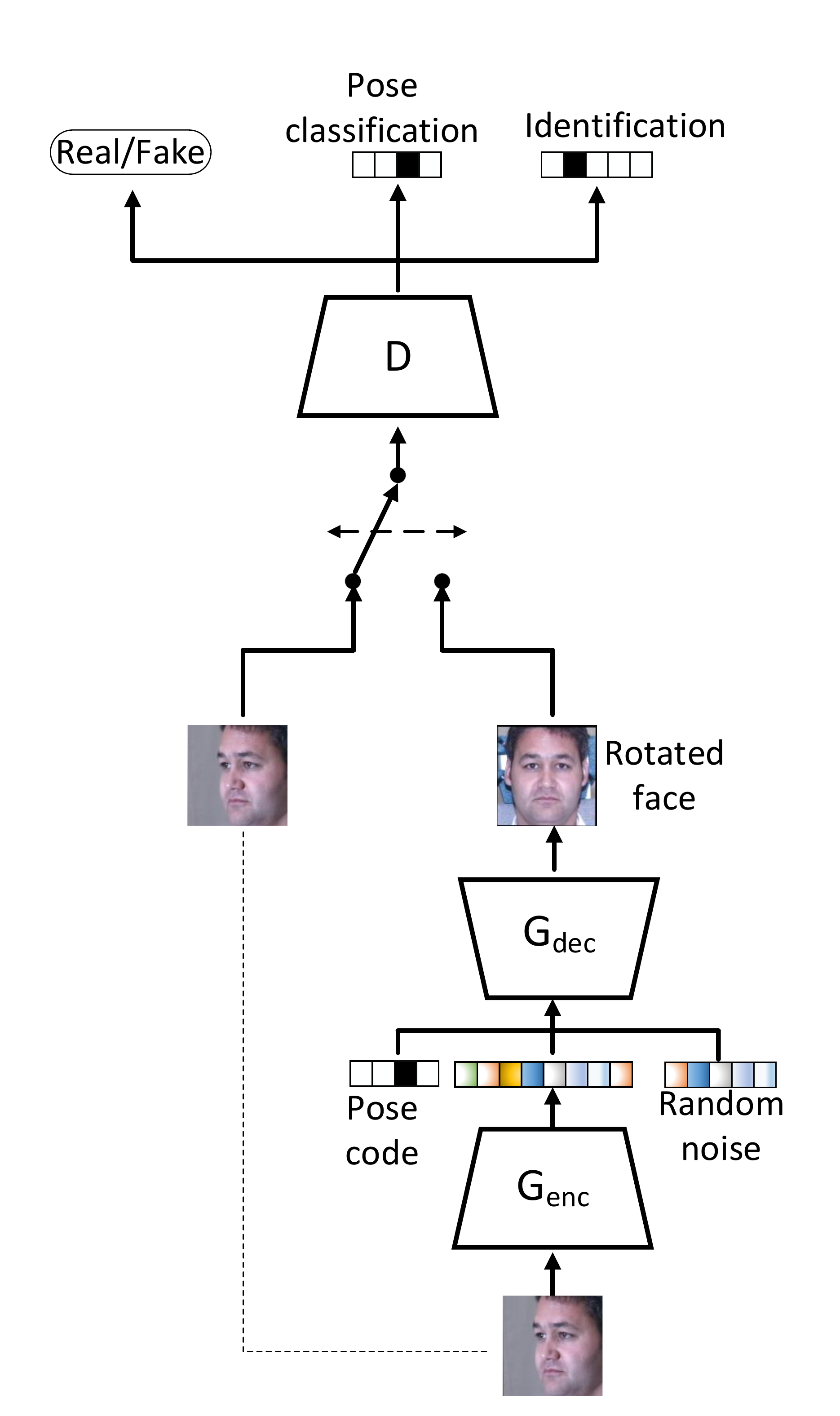}
}
\subfloat[DED-GAN]{
 \label{DED-GAN}
  \includegraphics[trim = 5mm 5mm 2mm 220mm, clip, height = .49\linewidth]{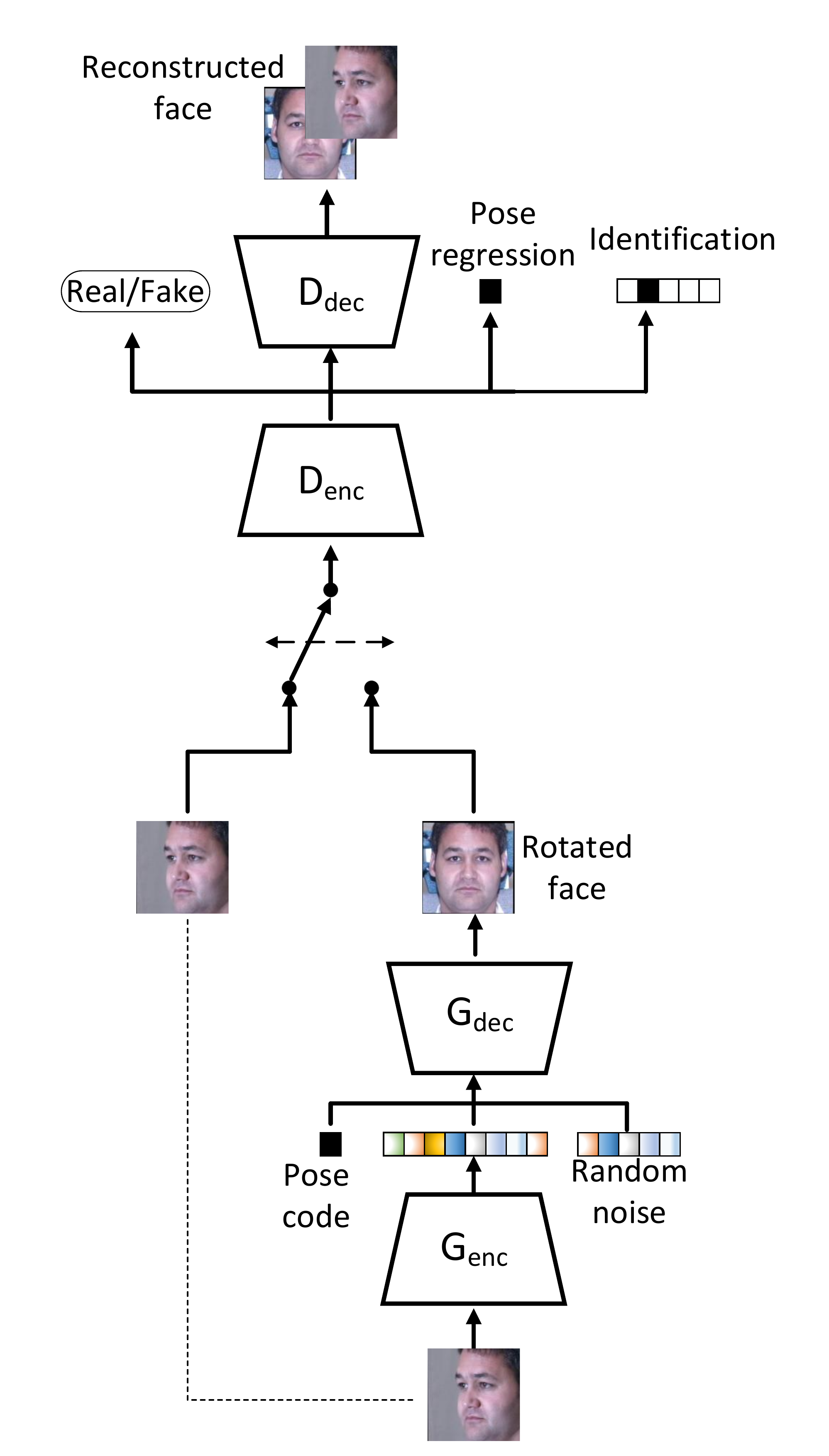}
}
\caption{Comparison of previous GANs architecture and our proposed DED-GAN.}
\label{Comparison with previous GANs}
\end{figure*}

Our Dual Encoder-Decoder based GAN (DED-GAN) model learns two tasks simultaneously: synthesis of different face poses, and pose-invariant face recognition. 
The encoder-decoder structured generator is used for face rotation and untangling the identity from pose variation. 
The encoder-decoder structured discriminator is used for facial reconstruction, pose estimation, identity classification, and real/fake adversarial learning. 
The architecture of our DED-GAN is shown in Fig.~\ref{DED-GAN}. 
We also show different architectures of earlier GANs such as Vanilla GAN, Auxiliary Classifier GAN and DR-GAN for comparison in Fig.~\ref{Vanilla GAN}, Fig.~\ref{Auxiliary Classifier GAN} and Fig.~\ref{DR-GAN}.
Fig.~\ref{DR-GAN}.
In contrast to DR-GAN, we add a decoder to the discriminator, which is optimised for pixel-wise loss defined in terms of the Wasserstein distance, to balance the generator and discriminator. 
We also code the pose using a continuous variate instead of the discrete variate commonly specified by a one-hot vector. 
As a result, the task of pose disentanglement in the discriminator can be formulated as one of pose regression instead of classification, which further benefits the learning process.

It should be noted that the Encoder-Decoder structured discriminator has also been successfully used in BEGAN~\cite{berthelot2017began}, with the aim of matching the pixel-wise loss distributions of reconstructed real and synthesised samples. 
Our method also incorporates an Encoder-Decoder as the backbone of the discriminator in order to achieve a balanced learning behaviour as part of weakly adversarial learning. 
Different from previous GANs, including BEGAN, our method not only sets out to match data distributions, but also attempts to match image reconstruction loss distributions. 
This is achieved by using a typical GAN objective combined with an additional equilibrium term. 
In order to provide a detailed description of our approach, we start by introducing the original GAN, followed by our proposed DED-GAN method.

\subsection{Generative Adversarial Network}
A typical GAN model consists of two networks pitted against one another in a two-player game: a generative model, $G$, is trained to synthesise images resembling the real data distribution and a discriminative model, $D$, is trained to distinguish the samples synthesised by $G$ and real ones from the training data. The generator generates unlabelled realistic samples from the latent variable model to improve the discriminative ability of the discriminator. To learn the generator's distribution $p_g$ over data $x$, we define a prior on input noise variables $p_z(z)$. The mapping $G(z; \theta_g)$ of $z$ into the data space is achieved by a neural network with parameters $\theta_g$, where $G$ is a differentiable function. A second neural network with parameters $\theta_d$ is defined by $D(x; \theta_d)$ that outputs a single scalar. $D(x)$ represents the probability that $x$ comes from the real data, $p_d$, rather than $p_g$. We train $D$ to maximise the probability of assigning the correct label to both training examples and samples from $G$. We simultaneously train $G$ to minimise $log(1-D(G(z)))$. In other words, the generator and discriminator are fighting against each other, which can be formulated as:
\begin{equation}
\begin{split}
\min_{G}\max_{D}L=& E_{x\sim p_d(x)}[logD(x)]+\\
                        & E_{z\sim p_z(z)}[log(1-D(G(z)))],\\
\end{split}
\end{equation}
where $z$ denotes a random noise, typically sampling from a Gaussian normal distribution, $p_z$. $G(z)$ denotes a sample synthesised by the generator and $p_d$ denotes the distribution of real data. It is proved in the original GAN~\cite{goodfellow2014generative} that this minimax game has a global optimum when the distribution $p_g$ of the synthetic samples converges to the distribution $p_d$ of the real samples. At the beginning of training, the samples generated by $G$ are extremely poor and thus they are rejected by $D$ with high confidence. This minimax game theoretically has a global optimum for $p_g = p_d$. $G$ and $D$ are trained to alternatively optimise the following objectives:
\begin{equation}
\begin{split}
\max_{D}L=& E_{x\sim p_d(x)}[logD(x)]+\\
                        & E_{z\sim p_z(z)}[log(1-D(G(z)))],\\
\end{split}
\end{equation}
\begin{equation}
\begin{split}
\min_{G}L=& E_{z\sim p_z(z)}[log(1-D(G(z)))].\\
\end{split}
\end{equation}
After several steps of the optimisation process, the generator and discriminator will reach the point at which neither can improve because $p_g = p_d$. The discriminator is unable to differentiate between the two distributions, \textit{i.e.} $D(x) = 1/2$.

\subsection{Dual encoder-decoder based GAN}
Our DED-GAN explicitly disentangles face imaging factors to obtain an interpretable face representation for PIFR and for face synthesis across poses. 
The backbone of DED-GAN consists of an encoder-decoder based generator and encoder-decoder based discriminator, as depicted in Fig.~\ref{DED-GAN}. 
It learns the representation of a face by using the generator, where the encoded output of the generator is the identity-preserving representation. The representation is one part of the input to the decoder to synthesise various faces of the same subject with different attributes, \textit{i.e.}, by virtually rotating the facial pose code. We not only match the distribution of face images by using classical real/fake adversarial learning, but also the distributions of the reconstruction error of  samples reconstructed from the representation by using pixel-wise adversarial learning. As numerous variations manifest in face images such as pose, illumination, and expression influence the face recognition even more than changes in identity, it is desirable to prevent the generator from generating different facial representations for the same person with different face poses. In this work we focus on pose variations, and disentangle the pose information as an explicit variation. This facilitates learning a truly discriminative face representation. 

\subsubsection{Problem Formulation}
Our method aims to train a generative adversarial model conditioned on the real face image $x$ and specified pose code $c$. Given a face image $x$ with label $y = \{y^a, y^d, y^c\}$, where $y^a$, $y^d$ and $y^c$ represent the labels for real/fake, identity and pose. 
There are two tasks in our learning method: to learn a disentangled identity representation for PIFR and to synthesise faces across poses with different pose code $c$.

Different from the discriminator in the original GAN, our discriminator could be seen as a multi-task CNN consisting of four components: $D =[D^a, D^d, D^c, D^r]$, where $D^a\in \mathbb{R}^1$ is for classical real/fake adversarial learning, $D^d\in \mathbb{R}^{N^d}$ is for identity classification with $N^d$ as the total number of subjects in the training set, $D^r\in \mathbb{R}^{N^c*N^w*N^h}$ is for face reconstruction and $D^c\in \mathbb{R}^{N^1}$ is for pose regression.

For the pose regression task, we first obtain the pose coefficients of all the training images.
To obtain the pose of an image, we use the MTCNN method to extract 5 facial landmarks for each face image~\cite{zhang2016joint}.
Then we transform face landmarks to the pose code using a statistical shape model~\cite{matthews2004active}. 
Mathematically, we can express the face shape with a base shape $s_0$ plus a linear combination of $n$ shape eigenvectors $s_i$ as:
\begin{equation}
s=s_0+\sum_{i=1}^n c_i s_i,
\end{equation}
where $s_0$ is the mean shape, $s_i$ is the $i$th shape eigenvector by applying principal component analysis to all the training shapes and $c_i$ is the corresponding coefficient.
In general, the first shape eigenvector controls pose variations of the model thus we use $c_1$ as the pose code $c$.

The discriminator aims to classify the face image $x$ as real or fake, to maximise the gap between the reconstruction error of real image and that of the synthetic image, and to estimate its identity and pose. Given an input image $x$, a random pose code $c$ and a random noise $z$, the generator $G$  generates a synthesised face image $G(x, c, z)$. The discriminator $D$ attempts to classify the image using the following objectives:
\begin{equation}
\begin{split}
L_{adv}^{D} = &E_{x,y\sim p_d(x,y)}[-logD^a(x)]+\\
& E_{x,y\sim p_d(x,y),z\sim p_z(z),c\sim p_c(c)}[-log(1-D^a(G(x,c,z)))],
\end{split}
\end{equation}

\begin{equation}
\begin{split}
L_{id}^D=E_{x,y\sim p_d(x,y)}[-logD_{y^d}^d(x)],
\end{split}
\end{equation}

\begin{equation}
\begin{split}
L_{pos}^D=E_{x,y\sim p_d(x,y)}|D_{y^c}^c(x)|,
\end{split}
\end{equation}

\begin{equation}
\begin{split}
L_{pixel}^D=E_{x\sim p_d(x),z\sim p_z(z),c\sim p_c(c)}|D^r(x)-k\cdot D^r(G(x,c,z))|.
\end{split}
\end{equation}
where $k$ is a trade-off parameter to balance the distribution of reconstruction error of real faces and that of synthetic faces. For clarity, we eliminate all subscripts for expected value notation, as all random variables are sampled from their respected distributions $(x,y)\sim p_d(x,y),z\sim p_z(z),c\sim p_c(c)$. $D^d$ is used for identity classification. It  should be noted that pose regression $D^c$ is used here rather than pose classification. The final objective for training $D$ is the weighted average of all objectives:
\begin{equation}
\begin{split}
min L^D=\lambda_aL_{adv}^D+\lambda_dL_{id}^D+\lambda_cL_{pos}^D+\lambda_rL_{pixel}^D,
\end{split}
\end{equation}
where $\lambda_a$, $\lambda_d$, $\lambda_c$ and $\lambda_r$ denote the weights of the four losses.

The generator $G$ consists of an encoder $G_{enc}$ and a decoder $G_{dec}$, where $G_{enc}$ aims to learn an identity-preserving representation $f(x)=G_{enc}(x)$ from a face image $x$,
$G_{dec}$ is tasked to synthesise a face image $G_{dec}(f(x), c, z)$ with identity $y^d$ and a target pose specified by $c$, and $z\in R^{N^z}$ is a noise variable, modelling other variations besides identity or pose. The pose code $c\in R^1$ is of continuous value. The goal of $G$ is to fool $D$ to classify $G(x, c, z)$ to the identity of input $x$ and estimate the target pose with the following objectives:
\begin{equation}
\begin{split}
L_{adv}^G = E_{x,y\sim p_d(x,y),z\sim p_z(z),c\sim p_c(c)}[-logD^a(G(x,c,z))],
\end{split}
\end{equation}

\begin{equation}
\begin{split}
L_{id}^G=E_{x,y\sim p_d(x,y)}[-logD_{y^d}^d(G(x,c,z))],
\end{split}
\end{equation}

\begin{equation}
\begin{split}
L_{pos}^G=E_{x,y\sim p_d(x,y)}|D_{y^c}^c(G(x,c,z))|,
\end{split}
\end{equation}

\begin{equation}
\begin{split}
L_{pixel}^G=E_{x\sim p_d(x),z\sim p_z(z),c\sim p_c(c)}|D^r(G(x,c,z))|.
\end{split}
\end{equation}
Similarly, the final objective for training the generator $G$ is the weighted average of each objective:
\begin{equation}
\begin{split}
min L^G=\mu_a L_{adv}^G+\mu_d L_{id}^G+\mu_c L_{pos}^G+\mu_r L_{pixel}^G.
\end{split}
\end{equation}
where $\mu_a$, $\mu_d$, $\mu_c$ and $\mu_r$ denote the weights of the four losses. 

\subsubsection{Pixel-wise loss}
While classical GANs try to match data distributions directly with $L_{adv}$, our method additionally aims to  match auto-encoder loss distributions using a pixel-wise loss $L_{pixel}$ based on Wasserstein distance. Firstly, we introduce the auto-encoder loss, and then we compute a lower bound to the Wasserstein distance between the auto-encoder loss distributions of real and generated samples.

Let $L:R^{N_x}\mapsto R^+$, denote the loss for training a pixel-wise auto-encoder defined as:
\begin{equation}
L(x)=|x-D(x)|^\eta 
\end{equation}
where $D: R^{N_x}\mapsto R^{N_x}$ is the auto-encoder, $\eta\in\{1,2\}$ is the target norm, and $x\in R^{N_x}$ is a sample of dimension $N_x$. Furthermore,  
let $\mu_{1,2}$ be two distributions of auto-encoder losses, and  $\Gamma(\mu_1,\mu_2)$ be the set all of couplings of $\mu_1$ and $\mu_2$, whose respective means are $m_{1,2}\in R$. The Wasserstein distance can be expressed as:
\begin{equation}
W_1(\mu_1,\mu_2)= \inf_{\gamma\in\Gamma(\mu_1,\mu_2)}E_{(x_1,x_2)\sim\gamma}[|x_1,x_2|]
\end{equation}
Using Jensen's inequality, we can derive a lower bound to $W_1(\mu_1,\mu_2)$:
\begin{equation}
\inf E[|x_1,x_2|]\geq \inf|E[x_1-x_2]|=|m_1-m_2|
\end{equation}
We design the discriminator to maximise $|m_1-m_2|$ by forcing $m_1\to 0, m_2\to \infty$. 
Given the discriminator and generator parameters $\theta_D$ and $\theta_G$, each to be updated by minimising the losses $L_{pixel}^D$ and $L_{pixel}^G$, we express the optimisation problem in terms of a pixel-wise loss function:
\begin{equation}
L_{pixel}^D= L(x)-k_t\cdot L(G(x))
\end{equation}
\begin{equation}
L_{pixel}^G= L(G(x))
\end{equation}
\begin{equation}
k_{t+1}=k_t+\lambda_k(\beta L(x)-L(G(x)))
\end{equation}
where $k_t$ controls how much emphasis is put on $L(G(x))$ during
gradient descent, $\lambda_k$ is the learning rate for $k$. $\beta$ is diversity ratio as a hyper-parameter to balance $L(x)$ and $L(G(x))$.

\section{Implementation Details}
\label{sec_implementation}
The proposed Dual Encoder-Decoder based GAN (DED-GAN) is composed of a generator $G$ and a discriminator $D$. Both are based on deep encoder-decoder networks. We follow the design for making $G$ in the DR-GAN. The modified CASIA Net~\cite{yi2014learning} is used as the backbone network. It consists of five convolution blocks, including one double-convolution block and four triple-convolution blocks, followed by an average pooling (AvePool) layer for feature extraction.

The generator $G$ is composed of an encoder $G_{enc}$ and a decoder $G_{dec}$, \textit{i.e.}, $G=[G_{enc}, G_{dec}]$. Given a face image $x$, the encoder's output code $e=G_{enc}(x)\in R^{N_e}$ from the AvePool layer is concatenated with a pose code $c\in R^{N_c}$ and a noise $z\in R^{N_z}$ to form $[e, c, z]$, which is used as the input of $G_{dec}$. $G_{dec}$ is a de-convolution neural network that transforms $[e, c, z]$ to a decoded face image, \textit{i.e.}, $\hat{x}=G_{dec}([e, c, z])$. $D_a$ and $D_r$ are used to force the distributions of both synthesised samples and their auto-encoder losses to match those of real samples.

The discriminator $D$ is composed of an encoder $D_{enc}$ and a decoder $D_{dec}$, \textit{i.e.}, $D=[D_{enc}, D_{dec}]$. Same as the generator, the backbone of the discriminator is also an encoder-decoder network where face reconstruction is $D_r$, aiming to increase the divergence of the auto-encoder loss distributions between real and synthesised samples. The code layer of the auto-encoder is followed by $D_a, D_c$ and $D_d$ where $D_a(x)$ is for real-fake classification, $D_c(x)$ is for pose regression and $D_d$ is for identity prediction. In Algorithm~\ref{algorithm_1}, we summarise the learning procedure of the proposed DED-GAN model. We use the Adam optimiser~\cite{kingma2014adam} for network training.
\begin{algorithm}
\label{algorithm_1}
\caption{The DED-GAN training algorithm}
\begin{algorithmic}
\STATE {\textbf{Input:}Training dataset $X$ and label $Y$. $X=\{x_1,x_2,...,x_N\}$. $Y$ includes the pose label and identity label: $Y=\{(y^{pos}_1,y^{id}_1),(y^{pos}_2,y^{id}_2),...,(y^{pos}_N,y^{id}_N)\}$. Initialise all the parameters $\theta=\{\theta_{g},\theta_{d}\}$ in generator and discriminator, trade-off hyper-parameters $\lambda_a, \lambda_d, \lambda_c, \lambda_r, \mu_a, \mu_d, \mu_c, \mu_r$ and Adam hyper-parameter $\alpha$. The number of iteration t$\leftarrow$ 0.\\
\textbf{Output:} $\theta=\{\theta_{g}, \theta_{d}\}$}
\STATE 1: \textbf{while} $\theta_g$ does not converge do.
\STATE 2: t$\leftarrow$ t+1.
\STATE 3: Sample noisy data $Z$ and pose code $C$ and compute the cost of $L^t (D)$ by $L^t (D) \leftarrow \lambda_a L^{{D}^t}_{adv}(X)+\lambda_d L^{{D}^t}_{id}(X)+\lambda_c L^{{D}^t}_{pos}(X)+\lambda_r L^{{D}^t}_{pixel}(X,Z,C)$ using equations(5)-(9).
\STATE 4: Compute the back propagation error to optimise discriminator $\Theta_d^{t} \leftarrow Adam(\nabla_{\theta_d^t}L^t (D),\alpha)$.
\STATE 5: Sample noisy data $Z$ and pose code $C$ and generate data $X_{g}=\{G(x_1,z_1,c_1),G(x_2,z_2,c_2),...,G(x_N,z_N,c_N)\}$.
\STATE 6: Compute the cost of $L^t (G)$ by $L^t (G) \leftarrow \mu_a L^{{G}^t}_{adv}(X)+\mu_d L^{{G}^t}_{id}(X)+\mu_c L^{{G}^t}_{pos}(X)+\mu_r L^{{G}^t}_{pixel}(X,Z,C)$ using equations(10)-(14).
\STATE 7: Fix the discriminator parameter $\Theta_d^{t}$ and compute the back propagation error to optimise generator $\Theta_g^{t} \leftarrow Adam(\nabla_{\theta_g^t}L^t (G),\alpha)$.
\STATE 8: \textbf{end while}
\end{algorithmic}
\end{algorithm}

All the experiments were preformed with the following settings. All face images were aligned to a canonical view of $100\times100$ in size. Randomly sampled regions of size $96 \times 96$ pixels selected from  $96\times96$ each  aligned face were cropped for data augmentation.
The image intensity was linearly scaled to the range of [-1,1]. All weights in the networks were initialised by a normal distribution with 0 mean and standard deviation 0.02.
We set the diversity ratio,  $\beta$, to 0.9. $k_t\in [0,1]$ controls how much emphasis is put on $L(G(x))$ during the network optimisation. We initialise $k_0=0$ and update $k$ in each training step. $\lambda_k$ is the learning rate for k. We set $\lambda_k$ to 0.001 in our experiments. We define the trade-off between the respective components of the loss function by setting $\lambda_a=1$, $\lambda_d=1$, $\lambda_c=0.1$, $\lambda_r=10$, $\mu_a=1$, $\mu_d=1$, $\mu_c=0.1$ and $\mu_r=10$ through numerous experiments. All experiments were run on a NVIDIA GeForce GTX Titan Xp card with CUDA 8.0 and cuDNN 6.0, implemented in Pytorch.

\section{Experiments}
\label{sec_experiment}
\subsection{Experimental Settings and Datasets}
We evaluate DED-GAN qualitatively and quantitatively under both constrained and unconstrained scenarios for face synthesis across poses and PIFR. 
Our models were trained separately on the Multi-PIE~\cite{gross2010multi} and CASIA~\cite{yi2014learning} datasets. 
For the qualitative evaluation, we show visualised results of face synthesis on Multi-PIE, CASIA, and CFP~\cite{sengupta2016frontal}.
For the quantitative evaluation, we measure face recognition performance using the learned facial representations with a cosine distance metric on the Multi-PIE, CFP, and LFW~\cite{huang2008labeled} datasets.
\begin{table*}
\begin{center}
\caption{DED-GAN and its partial variants performance comparison.}
\begin{tabular}{ccccccc}
\hline 
Model & $0^\circ$  & $\pm15^\circ$ & $\pm30^\circ$ & $\pm45^\circ$ & $\pm60^\circ$ & Average\\
\hline 
\hline 
DED-GAN(-$D^c$) &99.62& 98.20& 95.78 & 92.04 & 86.11  & 93.47   \\
DED-GAN(-$D^r$) &99.33 &98.62  & 96.86 & 92.39 & 86.20 &93.92   \\
DED-GAN(-$D^a$) &99.48 & 99.04 &97.47  &93.47  & 85.65 &94.36  \\
DED-GAN(using pose classification) &99.72  &99.15  &97.76  &94.12  &84.96  &94.64  \\
DED-GAN* & \textbf{99.95} & \textbf{99.45} & \textbf{98.02} & \textbf{94.88} & \textbf{87.82} &  \textbf{95.75}\\ 
\hline 
\end{tabular} 
\end{center}
\end{table*}

The \textbf{Multi-PIE} dataset is the largest mult-iview face recognition benchmark in the constrained scenario. 
It contains more than 750,000 images of 337 identities recorded during five months. Each identity has images captured under 15 poses and 20 illuminations. These images were captured in four sessions during different periods. Like the previous methods, we evaluate our algorithm on a subset of the Multi-PIE database, where each identity has images from all the four sessions under nine poses from yaw angles $-60^\circ$ to $+60^\circ$. For a fair comparison, we follow the setting used in DR-GAN~\cite{tran2018representation}. 
We evaluate our method on the Multi-PIE dataset setting 2. The first 200 subjects are used for training and the remaining 137 subjects are used for testing. Different from DR-GAN in which the supervised pose information is used, we use MTCNN to extract five landmarks and then transform the landmarks to a pose label. In testing, one frontal view with neural illumination is used as the gallery image and other images are used as probes. Therefore, we have $N^d=200$ for identity classification, $N^p=1$ for pose regression, $N^a=1$ for real/fake classification and $N^r=3\times96\times96$ for colour image reconstruction. We set the dimension of the embedding feature and uncompressed noise to $N^f=320$, and $N^z=50$ respectively.

The \textbf{CASIA} dataset offers 494,414 in-the-wild face images of 10,575 subjects.
It is a widely used large-scale database for face recognition. We train our model on this dataset to evaluate the performance of our model on a realistic dataset. We have $N^d=10,575$, $N^p=1$. $N^f$ and $N^z$ are set as for Multi-PIE. We also evaluate the performance of our model in terms of quality of synthesised face poses. 

\textbf{CFP} contains 7,000 images of 500 subjects, where each subject has 10 frontal and 4 profile face images. The data are randomly organised into 10 splits, each containing an equal number of frontal to frontal and frontal to profile pairs, with 350 intra pairs and 350 non-matching pairs, respectively. We evaluate the face verification performance in terms of front-to-front and profile-to-front matching. We also evaluate the performance of our model on its ability to  synthesise faces across pose variations.

The \textbf{LFW} database contains 13,233 face images of 5,749 identities. The images were obtained by trawling the internet followed by face centring, scaling and cropping based on the bounding boxes provided by an automatic face detector. 
The LFW data have large in-the-wild variability, \textit{e.g.}, in-plane rotations, non-frontal poses, non-frontal illumination, varying expressions and so on. The verification set consists of 10 folders, each with 300 matching pairs and 300 non-matching pairs. We measure the face verification performance and compare it with existing methods.

\subsection{Ablation Study}
Our discriminator is designed as a multi-task CNN with four components, namely $D^a$, $D^c$, $D^d$ and $D^r$ for real/fake classification, pose regression, identification and face reconstruction respectively. While $D^d$ surely plays a significant role in assisting the model to preserve the face identity, it is instructive to understand the role of the remaining components. In this subsection, the effect of the four loss functions on the recognition performance is investigated. The results are presented in Table I which reports
the recognition performance of DED-GAN partial variants with each of $D$ components removed. While the variant without adversarial loss $D^a$ exhibits a slight performance drop, the models without face reconstruction $D^r$ and pose regression $D^c$ losses are degraded more severely. When removing $D^c$, there is no pose label to supervise the face discrimination, especially for the profile faces. The average accuracy of DED-GAN partial variants without pose estimation reduces from 95.75$\%$ to 93.47$\%$. This can be attributed to the pose information being entangled with identity in the feature representation. 

Table I also presents the performance of our model without face reconstruction $D^r$. The average accuracy drops from 95.75$\%$ to 93.92$\%$. This shows that facial reconstruction is almost equally important to pose estimation. This suggests that the encoder-decoder structured discriminator successfully  balances the training of the two players in GAN.
\begin{figure}
\begin{minipage}{0.48\linewidth}
  \includegraphics[width=8.8cm]{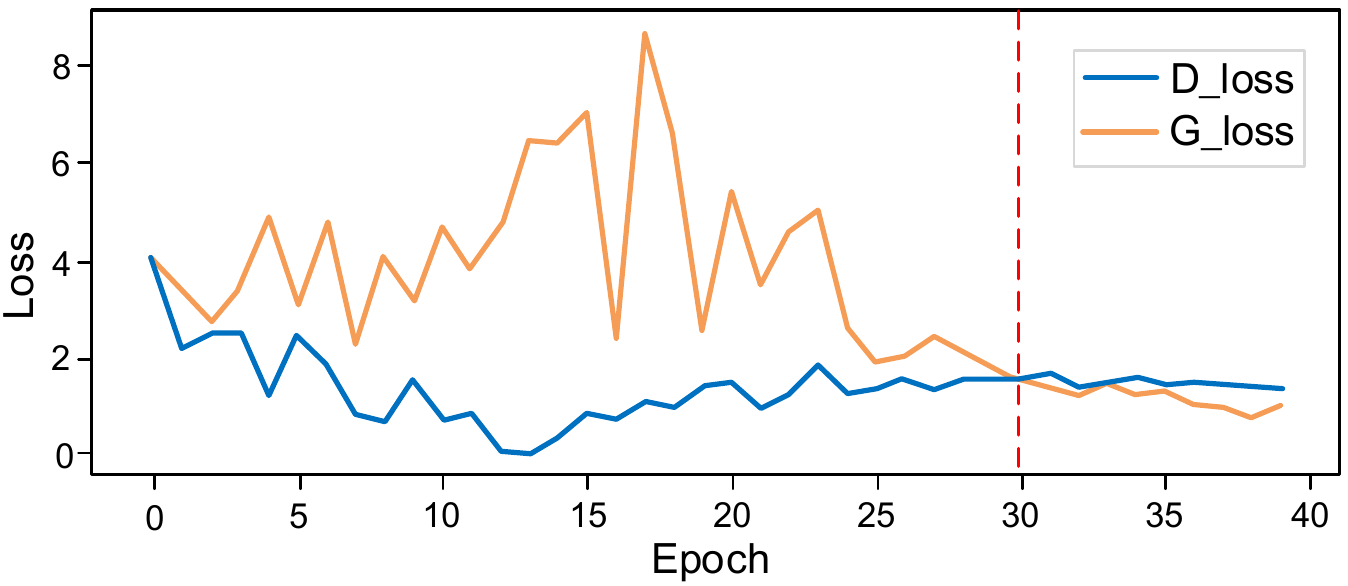}
\end{minipage}
\vfill
\begin{minipage}{.48\linewidth}
  \includegraphics[width=8.8cm]{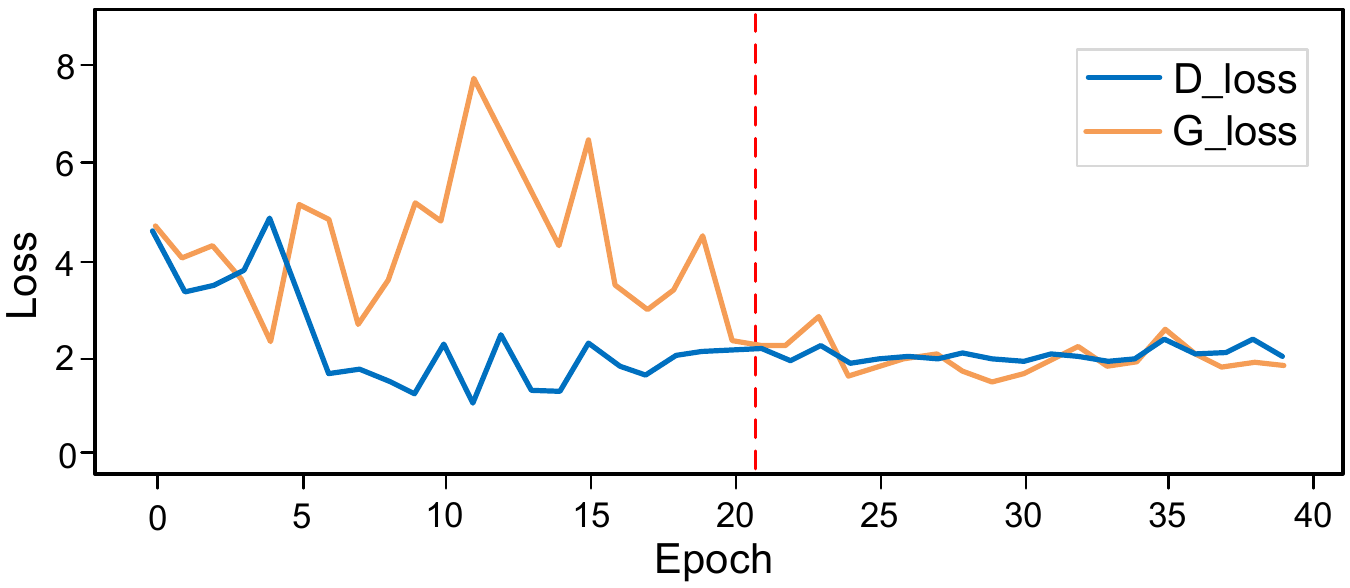}
\end{minipage}
\caption{Comparison of training loss of DED-GAN without(top) and with(bottom) pixel-wise loss on Multi-PIE (The top shows that training losses of generator and discriminator of DED-GAN without pixel-wise loss(DED-GAN($-D^r$)); The bottom shows that training losses of generator and discriminator of DED-GAN with pixel-wise loss(DED-GAN$^*$)).}
\end{figure}

To gauge the impact of using pose regression, rather than pose classification, we train separate DED-GAN models using the respective formulations. The results show that the performance of the model based on pose classification is lower by about 1\%. Thus continuous pose variation used for regression benefits for preserving more information about the pose.

The pixel-wise loss could effectively balance the generator and discriminator and get a fast convergence of training. To evaluate whether the pixel-wise loss could boost the convergence performance of DED-GAN, we compare the GAN loss with and without reconstruction task. Fig.~2 shows that DED-GAN without pixel-wise loss almost achieves convergence after 30 epochs. However, DED-GAN with pixel-wise loss gets a balance between generator and discriminator after about 20 epochs. The additional reconstruction task with pixel-wise loss suggests a fast and stable training manner between the generator and discriminator of GAN. We also compare the performance of DED-GAN with and without pixel-wise loss on the test accuracy and synthesised faces. As shown in Fig.~3, the DED-GAN with pixel-wise loss almost gets a stable test accuracy after 20 epochs training, while the DED-GAN without pixel-wise loss gets a stable accuracy at about 30 epochs. Fig.~4 shows the synthesised faces of DED-GAN with and without pixel-wise loss every five epochs during training. The result also shows that DED-GAN with pixel-wise loss could boost the quality of synthesised faces during training.
\begin{figure}
\centering
\includegraphics[width=8cm]{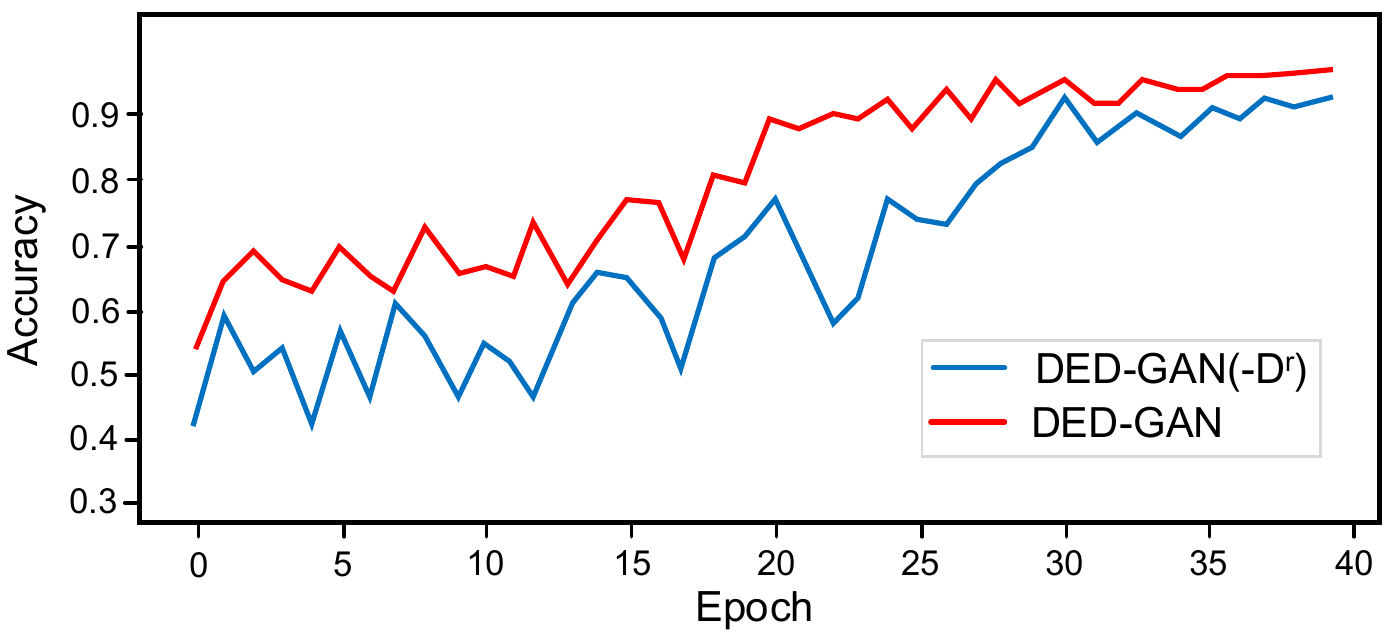}
\caption{Comparison of test accuracy of DED-GAN without (blue) and with (red) pixel-wise loss on Multi-PIE.}
\end{figure}

\begin{figure}
\centering
\includegraphics[width=8cm]{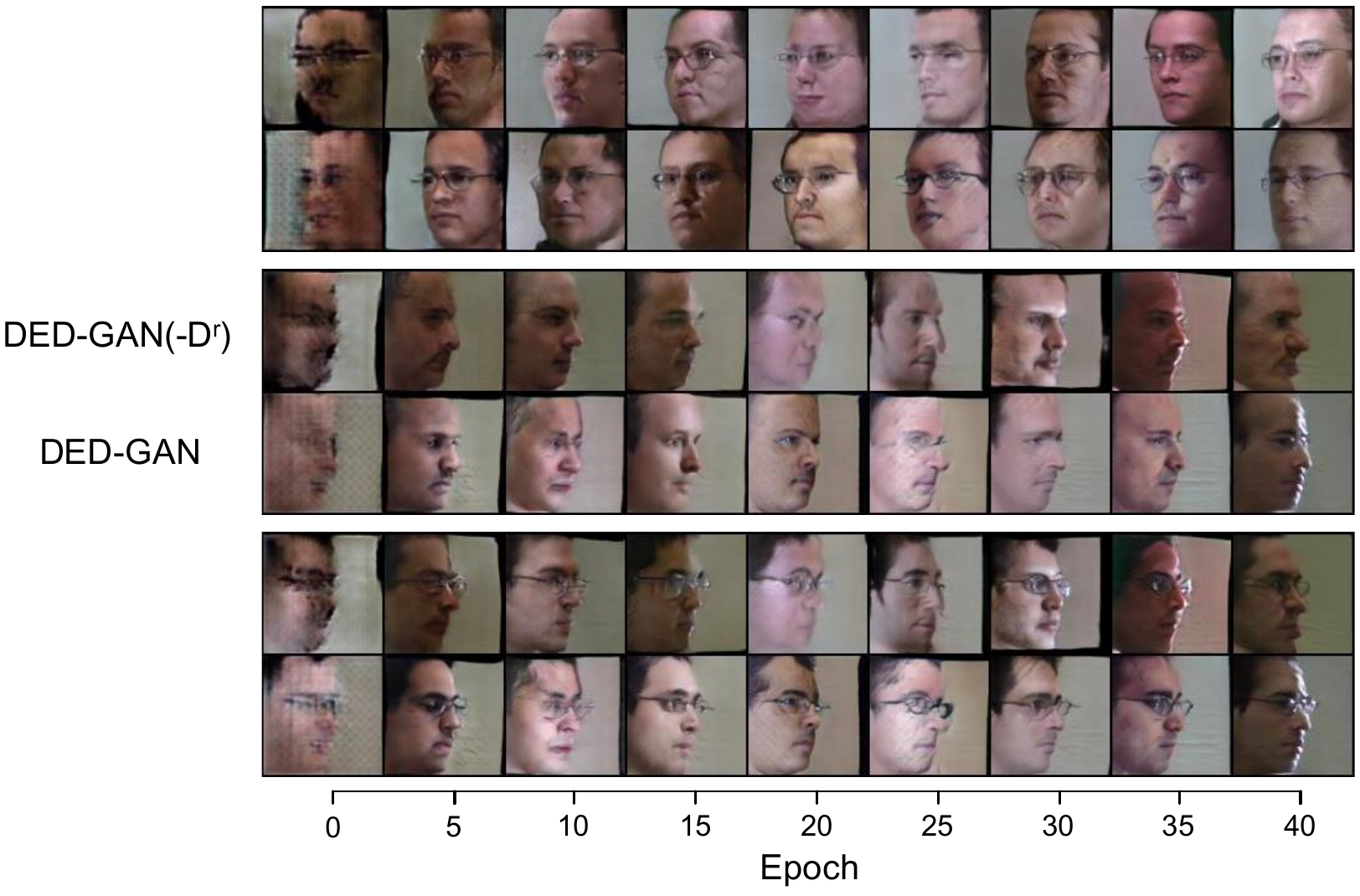}
\caption{Comparison of some synthesised faces of DED-GAN without (top) and with (bottom) pixel-wise loss on Multi-PIE.}
\end{figure}

\subsection{Face Synthesis}
To verify the performance of our method in terms of the quality of face synthesis across poses, a number of experiments are conducted on Multi-PIE, CASIA, and CFP datasets.
In the first experiment, we compared the synthesised faces with different poses between DR-GAN and our method on Multi-PIE. The synthesised faces are verified on the test set of the Setting 2. Hence, there is no overlap of subjects between the training and test datasets. Given a random input face, we generate synthesised faces within a pose range of $\pm 60^{\circ}$. The experimental results are shown in Fig.~5. We can see that the pose estimation capability helps to generate faces across poses and successfully disentangle pose variation from the feature vector in both methods. However, the quality of the faces synthesised by our method appears to be better than that of those output by the DR-GAN in texture, shape, as well as identity preserving characteristics.


\begin{figure}
\begin{minipage}{0.48\linewidth}
  \includegraphics[width=8.8cm]{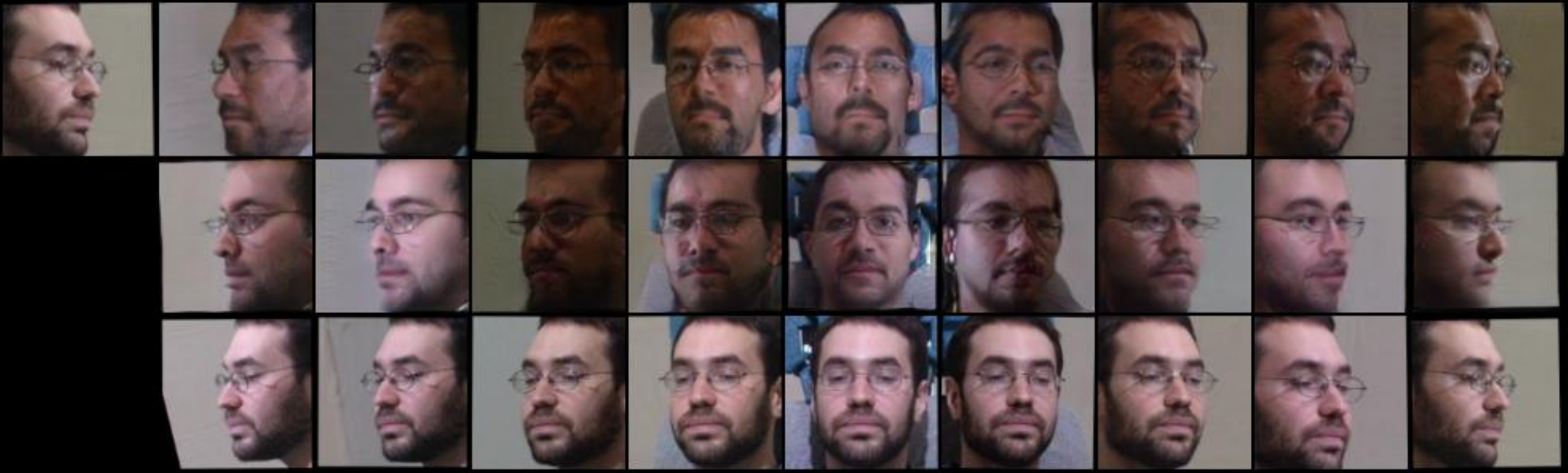}
\end{minipage}
\vfill
\begin{minipage}{.48\linewidth}
  \includegraphics[width=8.8cm]{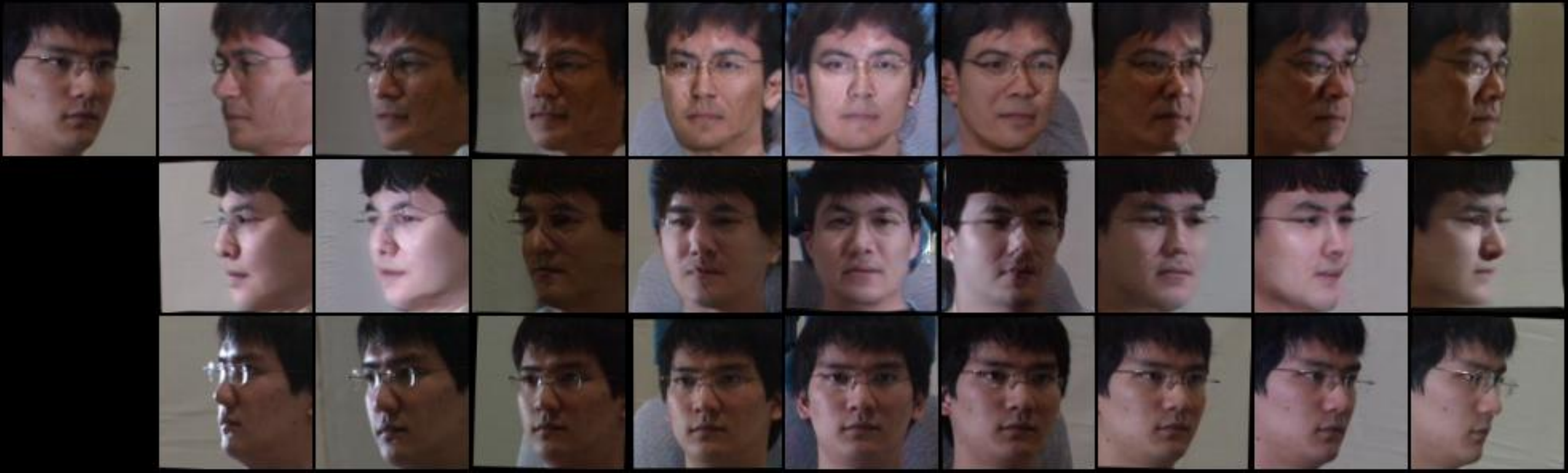}
\end{minipage}
\vfill
\begin{minipage}{0.48\linewidth}
  \includegraphics[width=8.8cm]{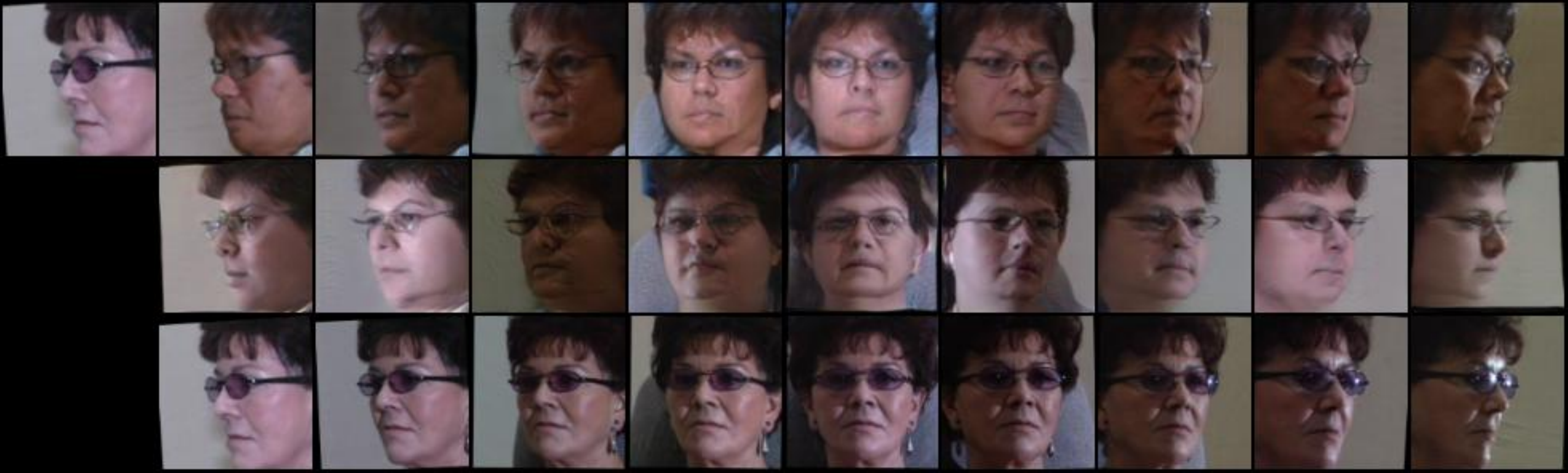}
\end{minipage}
\caption{Comparison of DR-GAN and DED-GAN generated images on Multi-PIE. Given three input images(the left column), the first ,fourth and seventh rows shows the faces synthesised by the DR-GAN; the second, fifth and eighth rows show the faces synthesised by DED-GAN; the third, sixth and ninth rows show the ground truth of nine poses within the degree from $-60^{\circ}$ to $60^{\circ}$.}
\end{figure}

For an objective evaluation of the relative quality of faces generated by the two types of GANs we use the Fr$\Acute{e}$chet Inception Distance(FID)~\cite{heusel2017gans}. For a feature function $\phi$ (by default, the Inception network's convolutional feature), FID models $\phi(p_d)$ and $\phi(p_g)$ as Gaussian random variables with empirical means $\mu_d$, $\mu_g$ and empirical covariance $\Sigma_d$, $\Sigma_g$, and computes $FID(p_d,p_g)=||\mu_d-\mu_g||+Tr(\Sigma_d+\Sigma_g-2(\Sigma_d\Sigma_g)^{1/2})$, which is the Fréchet distance between the two Gaussian distributions. Table II compares the FID scores between DR-GAN and DED-GAN. DED-GAN achieves a lower FID score than DR-GAN, which means that the faces synthesised by DED-GAN are more similar to real ones than those produced by DR-GAN.
\begin{table}
\begin{center}
\caption{Comparison of FID score.}
\begin{tabular}{cc}
\hline 
Model & FID score\\
\hline 
\hline 
DR-GAN & 71.25\\ 
DED-GAN & \textbf{57.03} \\ 
\hline 
\end{tabular} 
\end{center}
\end{table}
\begin{figure*}
\centering
\includegraphics[width=18cm]{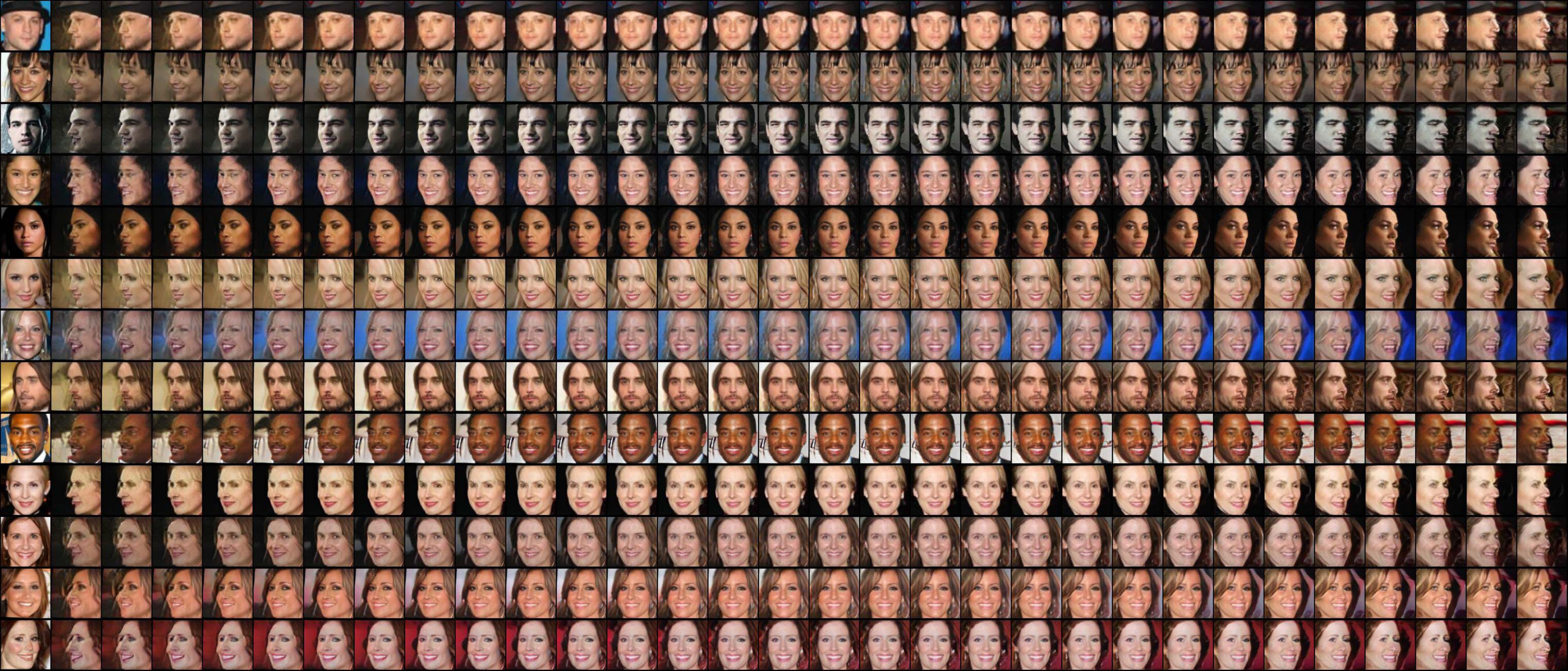}
\caption{Face manifold across poses on the CASIA database(Input is at the first column, the faces from the 2nd column to the last one are the manifold of synthesised faces with the same identity by changing the value of pose code from -17 to 17).}
\end{figure*}

To further demonstrate the ability to disentangle the pose generative factor from other face attributes, we also evaluate the performance of our model on face synthesis across poses on another two uncontrolled datasets CASIA and CFP. We use MTCNN to extract five facial landmarks for each face and then transform the landmarks to pose label by a statistical shape model. The CASIA facial distribution across poses is illustrated in Fig.~7 where the value zero denotes the frontal face. Note that, different from the previous methods, DED-GAN can rotate an input face to any pose controlled explicitly by the pose code. Hence, DED-GAN can synthesise both frontal and profile faces. Fig.~6 shows the pose manifold of generated faces by changing the value of the pose code. Every row denotes the faces with the same identity. The first column is the input face and the other columns show the manifold of the synthesised faces with a smoothly changing value of the pose code from -17 to 17. We can clearly see that our model well preserves the identity well as we change the pose code. It also shows that the pose variation is explicitly untangled from the other face attributes including identity.
\begin{figure}
\centering
\includegraphics[width=9cm]{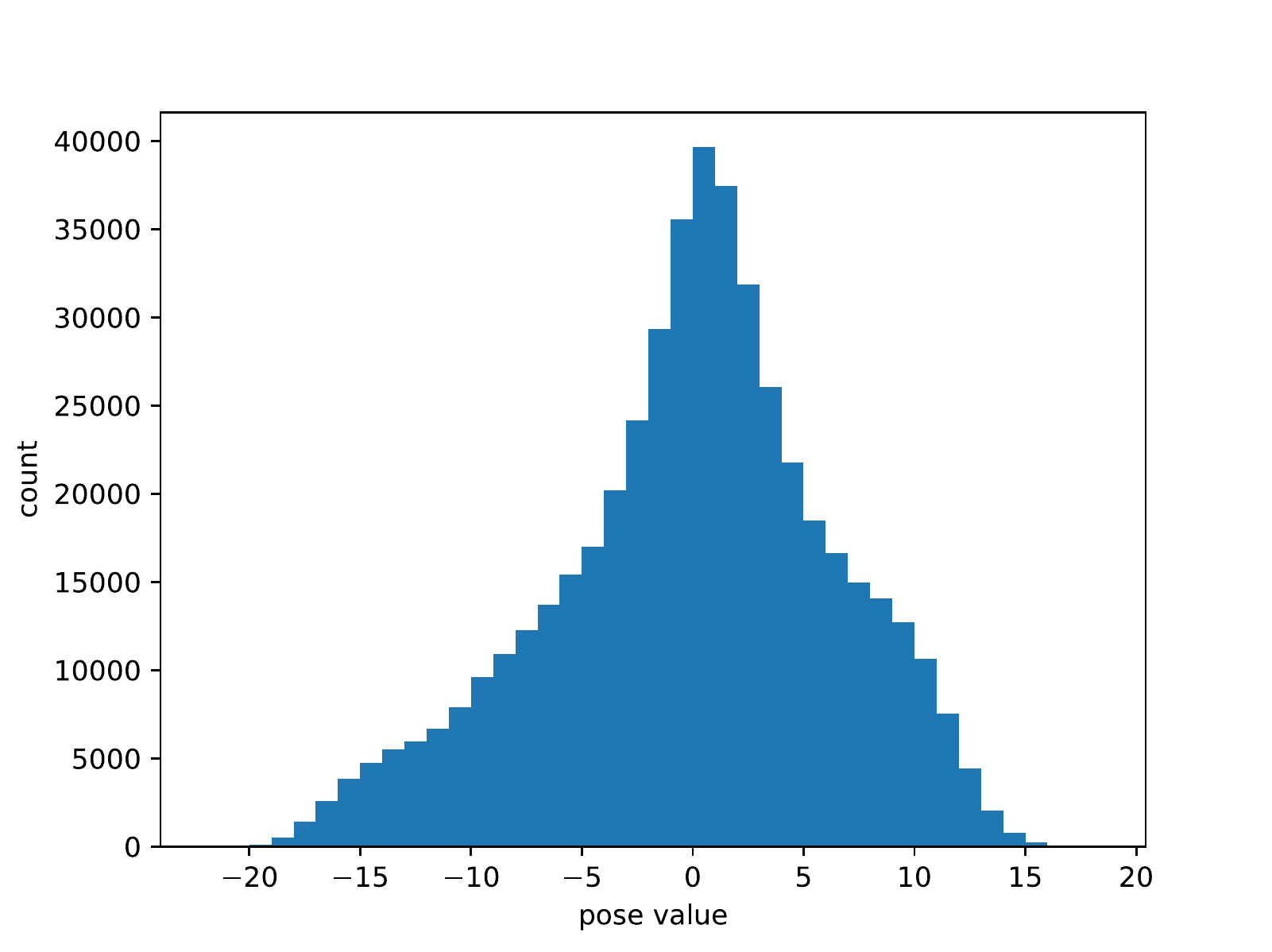}
\caption{Face distribution across poses on the CASIA database.}
\end{figure}

We also test the face frontalisation performance for unseen faces on the CFP dataset as shown in Fig.~8. Every column shows the faces of the same identity. Given an input profile face, we separately generate the frontal faces by DR-GAN and our method. The up and down rows show the input profile faces and paired real frontal faces separately. The second and third rows show the synthesised frontal faces by setting the pose code to zero. We can see that both methods can untangle the face representation from pose variation and generate  frontal faces. However, the faces synthesised by our method appear better in terms of texture detail and in preserving the face identity.
\begin{figure}
\centering
\includegraphics[width=8.5cm]{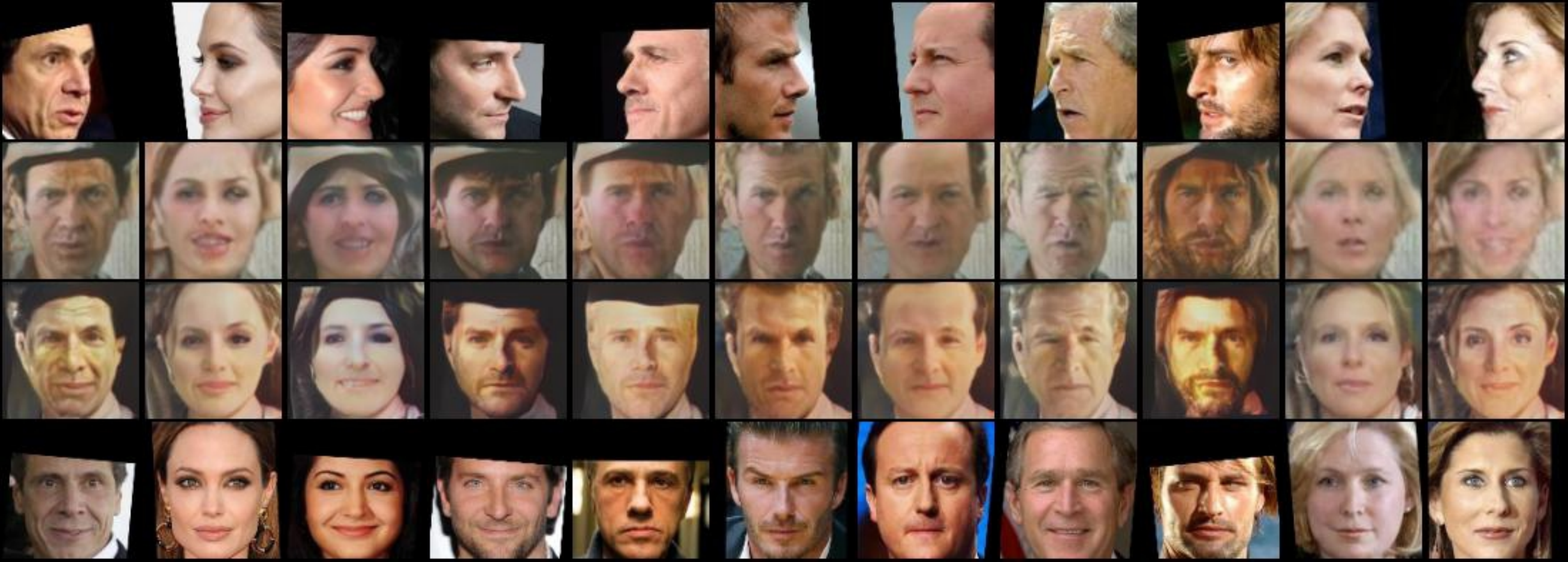}
\caption{Some face frontalisation results comparison on CFP database (From top to bottom: input images, DR-GAN frontalised faces, our frontalised faces, real frontal faces).}
\end{figure}

\begin{table*}
\begin{center}
\caption{Rank-1 recognition rates(\%) across views, illuminations and sessions under Multi-PIE.}
\begin{tabular}{ccccccc}
\hline 
Model & $0^\circ$  & $\pm15^\circ$ & $\pm30^\circ$ & $\pm45^\circ$ & $\pm60^\circ$ & Average\\
\hline 
\hline 
Zhu \textit{et al.}~\cite{zhu2014multi}& 95.70 & 92.80 & 83.70 & 72.90 & 60.10 &  79.30\\
Yim \textit{et al.}~\cite{yim2015rotating}& 99.50 & 95.00 & 88.50 & 79.90 & 61.90 &  83.30\\
DR-GAN~\cite{tran2017disentangled} & 97.00 & 94.00 & 90.10 & 86.20 & 83.20 &  89.20 \\
DR-GAN$_{am}$~\cite{tran2018representation} & 98.10 & 95.00 & 91.30 & 88.00 & 85.80 &  90.80 \\
FF-GAN~\cite{yin2017towards} & - & 94.60 & 92.50 & 89.70 & 85.20 & - \\
Light CNN~\cite{wu2018light} & - & 98.59 & 97.38 & 92.13 & 62.09 & - \\
\hline 
DED-GAN* & \textbf{99.95} & \textbf{99.45} & \textbf{98.02} & \textbf{94.88} & \textbf{87.82} &  \textbf{95.75}\\ 
\hline 
\end{tabular} 
\end{center}
\end{table*}

\subsection{Face Recognition}
One motivation for disentangled face representation learning is to see, whether the untangled representation helps to preserve the identity information, and thus boost the performance in face recognition. To verify this, we also show quantitative results obtained in PIFR experiments. We evaluate our method on Multi-PIE, CFP, and LFW for identification and verification tasks. The features are extracted from $G_{enc}$ in all the experiments. The cosine distance between two representations is used for face recognition in the test step.
\subsubsection{Face Identification on the Multi-PIE Database}
In the first experiment in PIFR, we evaluate the performance of DED-GAN on the Multi-PIE dataset. We compare our method with other state-of-the-art face recognition methods. Our model achieves the best accuracy in different pose categories, with the most significant improvement noted for  for the profile faces as shown in Table III. It shows that our method can remove the effects of pose, and retain the intrinsic face shape and structure information of identity.

\subsubsection{Face Verification on the CFP Database}
To further demonstrate the advantages of our method in PIFR, we evaluate it on an uncontrolled dataset. For in-the-wild setting, we train our model on CASIA and test it on the CFP database. 
Th experiments performed on the CFP dataset aim to compare the capacity of the face verification approaches across diverse poses. More specifically, the matching is performed between the frontal view (yaw angle $<10^\circ$) and profile view (yaw angle $>60^\circ$). The evaluation reports the mean and standard deviation of accuracy, over 10 splits, for both frontal to frontal and frontal to profile face verification settings.
The verification results are shown in Table IV. Our method again yields better verification performance on both frontal-frontal and frontal-profile matching sub-tasks. Thanks to the more stable training structure and more detailed pose information injected into our method, DED-GAN achieves about a one percent performance improvement over DR-GAN.

\begin{table}
\begin{center}
\caption{Face verification accuracy($\%$) comparison on on CFP.}
\begin{tabular}{ccc}
\hline 
Model & Frontal-Frontal  & Frontal-Profile\\
\hline 
\hline 
Sengupta \textit{et al.}~\cite{sengupta2016frontal}& $96.40\pm0.69$ & $84.91\pm1.82$\\
Sankarana \textit{et al.}~\cite{sankaranarayanan2016triplet}& $96.93\pm0.61$ & $89.17\pm2.35$\\
DR-GAN~\cite{tran2017disentangled} & $97.13\pm0.62$ & $90.82\pm0.28$ \\
Human & $96.24\pm0.67$ & $94.57\pm1.10$\\
\hline 
DED-GAN* & \textbf{97.99}$\pm$\textbf{0.85} & \textbf{91.58}$\pm$\textbf{1.38}\\
\hline 
\end{tabular} 
\end{center}
\end{table}

\subsubsection{Face Verification on the LFW Database}
To evaluate the performance on the in-the-wild dataset further, we test the  models described in the previous subsection on the LFW database.
Table V shows the accuracy achieved by different methods. As expected, our method DED-GAN delivers the best accuracy, namely  97.52$\%$, which is comparable with other state-of-the-art methods. Although DED-GAN is not trained on the LFW dataset, the untangled discriminative representation generalises to other datasets, including in-the-wild datasets.
\begin{table}
\begin{center}
\caption{Face verification accuracy($\%$) comparison on LFW.}
\begin{tabular}{cc}
\hline 
Model & Accuracy($\%$)\\
\hline 
\hline 
LFW-3D~\cite{hassner2015effective} & 93.62\\
LFW-HPEN~\cite{zhu2015high} & 96.25\\
FF-GAN~\cite{yin2017towards} & 96.42 \\
\hline 
DED-GAN* & \textbf{97.52}\\
\hline 
\end{tabular} 
\end{center}
\end{table}

\section{Conclusion}
\label{sec_conclusion}
We propose a new GAN-based pose-invariant model (DED-GAN) for disentangled representation learning to address the challenging problem of pose-invariant face recognition and photo-realistic face synthesis across poses. To the best of our knowledge, this is the first time that a dual encoder-decoder structured GAN has been used to learn disentangled face representation. The encoder-decoder structured generator is used for face rotation and learning disentangled face representation. The encoder-decoder structured discriminator is used for facial reconstruction and for predicting identity, as well as for estimating the pose. The Encoder-decoder structured discriminator with
the additional pixel-wise loss improves the training efficiency and stability of our GAN. A continuous pose encoding provides more detail pose information and benefits the discriminative representation by untangling the identity and pose. Extensive quantitative and qualitative experimental results show that our method is competitive compared to state-of-the-arts approaches to PIFR and to face synthesis across poses. In future, we plan to  incorporate more discriminative information into the design of DED-GAN by extending the network to deal explicitely  with other image generative factors, including illumination, expression, age, and occlusion. 

\section*{Acknowledgment}
This work is supported in part by the EPSRC Programme Grant (FACER2VM) EP/N007743/1, EPSRC/dstl/MURI
project EP/R018456/1, the National Natural Science Foundation of China \{61672265, U1836218, 61876072\}, the 111 Project of Chinese Ministry of Education \{B12018\} and the China Scholarship Council.

\ifCLASSOPTIONcaptionsoff
  \newpage
\fi



%



\bibliographystyle{IEEEtran}
\bibliography{bibliography}

%






\end{document}